\documentclass[final]{cvpr}

\usepackage{times}
\usepackage{epsfig}
\usepackage{graphicx}
\usepackage{amsmath}
\usepackage{amssymb}

\usepackage{booktabs}  
\usepackage{multirow}

\usepackage[pagebackref=true,breaklinks=true,colorlinks,bookmarks=false]{hyperref}

\def\cvprPaperID{7} 
\def\confYear{CVPR 2021}

\begin{document}

\title{Evaluating the Immediate Applicability of\\Pose Estimation for Sign Language Recognition}


\author{
Amit Moryossef\textsuperscript{1,2} \;\; 
Ioannis Tsochantaridis\textsuperscript{2} \;\; \\
Joe Dinn\textsuperscript{3} \;\;
Necati Cihan Camgöz\textsuperscript{3} \;\; 
Richard Bowden\textsuperscript{3} \;\; 
Tao Jiang\textsuperscript{3} \\
Annette Rios\textsuperscript{4} \;\; 
Mathias Müller\textsuperscript{4} \;\; 
Sarah Ebling\textsuperscript{4} \;\; 
\\
\texttt{amitmoryossef@gmail.com, ioannis@google.com} \\
\texttt{\{j.dinn, n.camgoz, r.bowden, t.jiang\}@surrey.ac.uk} \\
\texttt{\{rios, mmueller, ebling\}@cl.uzh.ch} \\

\textsuperscript{1}Bar Ilan University, \textsuperscript{2}Google \\
\textsuperscript{3}University of Surrey, \textsuperscript{4}University of Zurich}



\maketitle

\begin{abstract}
Sign languages are visual languages produced by the movement of the hands, face, and body. 
In this paper, we evaluate representations based on skeleton poses, as these are explainable, person-independent, privacy-preserving, low-dimensional representations. Basically, skeletal representations generalize over an individual's appearance and background, allowing us to focus on the recognition of motion. But how much information is lost by the skeletal representation? We perform two independent studies using two state-of-the-art pose estimation systems. We analyze the applicability of the pose estimation systems to sign language recognition by evaluating the failure cases of the recognition models. Importantly, this allows us to characterize the current limitations of skeletal pose estimation approaches in sign language recognition.
\end{abstract}

\section{Introduction}\label{sec:introduction}

Sign languages are visual languages produced by the movement of the hands, face, and body.
As languages that rely on visual communication, recordings are in video form.
Current state-of-the-art sign language processing systems rely on the video to model tasks such as sign language recognition (SLR) and sign language translation (SLT). However, using the raw video signal is computationally expensive and can lead to overfitting and person dependence. 

In an attempt to abstract over the video information, skeleton poses have been suggested as an explainable, person-independent, privacy-preserving, and low-dimensional representation that provides the signer body pose and information on how it changes over time.
Theoretically, skeletal poses contain all the relevant information required to understand signs produced in videos, except for interactions with elements in space (for example, a mug or a table).

The recording of accurate human skeleton poses is difficult and often intrusive, requiring signers to wear specialized and expensive motion capture hardware. Fortunately, advances in computer vision now allow the estimation of human skeleton poses directly from videos.
However, as these estimation systems were not specifically designed with sign language in mind, we currently do not understand their suitability for use in processing sign languages both in recognition or production. 

In this paper, we evaluate two pose estimation systems and demonstrate their suitability (and limitations) for SLR by conducting two independent studies on the CVPR21 ChaLearn challenge \cite{Sincan:CVPRW:2021}.
Because we perform no pretraining of the skeletal model, the final results are considerably lower than potential end-to-end approaches (\S\ref{sec:experiments}). The results demonstrate that the skeletal representation loses considerable information. To better understand why, we evaluate our approaches (\S\ref{sec:results}), categorize their failure cases (\S\ref{sec:analysis}), and conclude by characterizing the attributes a pose estimation system should have to be applicable for SLR (\S\ref{sec:conclusions}).

\section{Background}\label{sec:background}
\subsection{Pose Estimation}

Pose estimation is the task of detecting human figures in images and videos to determine where various joints are present in an image. This area has been thoroughly researched
\cite{pose:pishchulin2012articulated, pose:chen2017adversarial, pose:cao2018openpose, pose:alp2018densepose, mediapipe2020holistic}
with objectives varying from the predicting of 2D/3D poses to a selection of
a small specific set of landmarks or a dense mesh of a person. Vogler \cite{vogler2005analysis} showed that the face pose correlates with facial non-manual features.

OpenPose \cite{pose:cao2018openpose, pose:simon2017hand, pose:cao2017realtime, pose:wei2016cpm} was the first multi-person system to jointly detect human body, hand, facial, and foot keypoints (135 keypoints in total) in 2D on single images. 
While this model can estimate the full pose directly from an image in a single inference, a pipeline approach is also suggested where
first,  the body pose is estimated and then independently  the hands and face pose by acquiring higher-resolution crops around those areas. 
Building on the slow pipeline approach, a single-network whole-body OpenPose model has been proposed \cite{pose:hidalgo2019singlenetwork}, 
which is faster and more accurate for the case of obtaining all keypoints.
Additionally, with multiple recording angles, OpenPose also offers keypoint triangulation to reconstruct the pose in 3D.

DensePose \cite{pose:alp2018densepose} takes a different approach.
Instead of classifying for every keypoint which pixel is most likely, similar to semantic segmentation, each pixel is classified as belonging to a body part. 
Then, for each pixel, knowing the body part, the system predicts where that pixel is on the body part relative to a 2D projection of a representative body model. 
This approach results in reconstructing the full-body mesh and allows sampling to find specific keypoints similar to OpenPose.

MediaPipe Holistic \cite{mediapipe2020holistic} attempts to solve the 3D pose estimation problem directly by taking a similar approach to
OpenPose, having a pipeline system to estimate the body and then the face and hands. 
It uses a dense mesh model for the face pose containing 468 points, but resorts to skeletal joints for the body and hands. Unlike OpenPose, the poses are estimated using regression rather than classification and are estimated in 3D rather than 2D.

\subsection{Sign Language Recognition}

Sign language recognition (SLR) is the task of recognizing a sign or a sequence of signs from a video.
This task has been attempted both with computer vision models, assuming the input is the raw video, and with poses, assuming the video has been processed with a pose estimation tool.

\subsubsection{Video to Sign}
Camgöz et al. \cite{cihan2018neural} formulate this problem as one of translation. 
They encode each video frame using AlexNet \cite{krizhevsky2012imagenet}, initialized using weights that were trained on ImageNet \cite{deng2009imagenet}. 
Then they apply a GRU encoder-decoder architecture with Luong Attention \cite{luong2015effective} to generate the signs. 
In a follow-up work \cite{camgoz2020sign}, they use a transformer encoder \cite{vaswani2017attention} to replace the GRU and use Connectionist Temporal Classification (CTC) \cite{graves2006connectionist} to decode the signs. 
They show a slight improvement with this approach over the previous one.

Adaloglou et al. \cite{adaloglou2020comprehensive} perform a comparative experimental assessment of computer vision-based methods for the SLR task.
They implement various approaches from previous research \cite{camgoz2017subunets, cui2019deep, dataset:joze2018ms} and test
them on multiple datasets \cite{dataset:huang2018video, cihan2018neural, dataset:von2007towards, dataset:joze2018ms} either for isolated sign recognition or continuous sign recognition.
They conclude that 3D convolutional models outperform models using only recurrent networks because they better capture temporal information and that convolutional models are more scalable given the restricted receptive field, which results from their ``sliding window'' technique.

\subsubsection{Pose to Sign}

Upper body poses have been widely used as a feature for computational sign language research \cite{cooper2011sign}, due to their signer-invariant representation capabilities. They have been included into recognition \cite{fang2017deepasl}, translation \cite{camgoz2020multi}, or detection \cite{detection:moryossef2020real} frameworks, either in raw coordinate form or as linguistically meaningful symbols extracted from joint coordinates \cite{cooper2010sign}.

Before the deep learning era, most sign language systems utilized specialized sensors, such as Kinect \cite{zafrulla2011american, chai2013sign}, to estimate signers pose in real-time \cite{Shotton2011a}. There have also been attempts to train models on sign language data \cite{pfister2012automatic, charles2014automatic, luzardo2013head} which extract low-resolution skeletons, i.e., few joints. However, these approaches suffered from noisy estimations and had deficient hand joint resolution.

As with any subfield of computer vision, human pose estimation also improved with the introduction of deep learning-based approaches. Open source, general-purpose human pose estimation models, such as convolutional pose machines \cite {wei2016cpm} and their predecessor OpenPose \cite{pose:cao2018openpose}, became widely used in sign language research. Ko et al. \cite{ko2019neural} utilized a transformer-based translation based purely on skeletal information. Albanie et al. \cite{albanie20_bsl1k} proposed using pose estimates to recognize co-articulated signs. They further used the pose estimates to train knowledge distillation networks and learn meaningful representations for downstream tasks.

\section{Experiments}\label{sec:experiments}
To evaluate whether pose estimation models are applicable for SLR, we participated in the CVPR21 ChaLearn challenge for person-independent isolated SLR on the Ankara University Turkish Sign Language (AUTSL) \cite{dataset:sincan2020autsl} dataset.
Even though the dataset includes Kinect pose estimations, Kinect poses have not been made available for the challenge.
We processed the dataset using two pose estimation tools: 1. OpenPose Single-Network Whole-Body Pose Estimation \cite{pose:hidalgo2019singlenetwork}; and 2. MediaPipe Holistic \cite{mediapipe2020holistic}; and made the data available via an open-source sign language datasets repository \cite{moryossef2021datasets}.

We approach the recognition task with two independent experiments performed by different teams unaware of the other team's work throughout the validation stage.
In the validation stage, each team focussed on one pose estimation approach, and in the test stage, both teams got access to both pose estimation outputs. We eventually submitted three systems: 1. based on \emph{OpenPose} poses; 2. based on \emph{Holistic} poses; 3. based on both \emph{OpenPose} and \emph{Holistic} poses combined (concatenated).

\subsection{Team 1}

\begin{figure}[b]
    \centering
    \includegraphics[width=\linewidth]{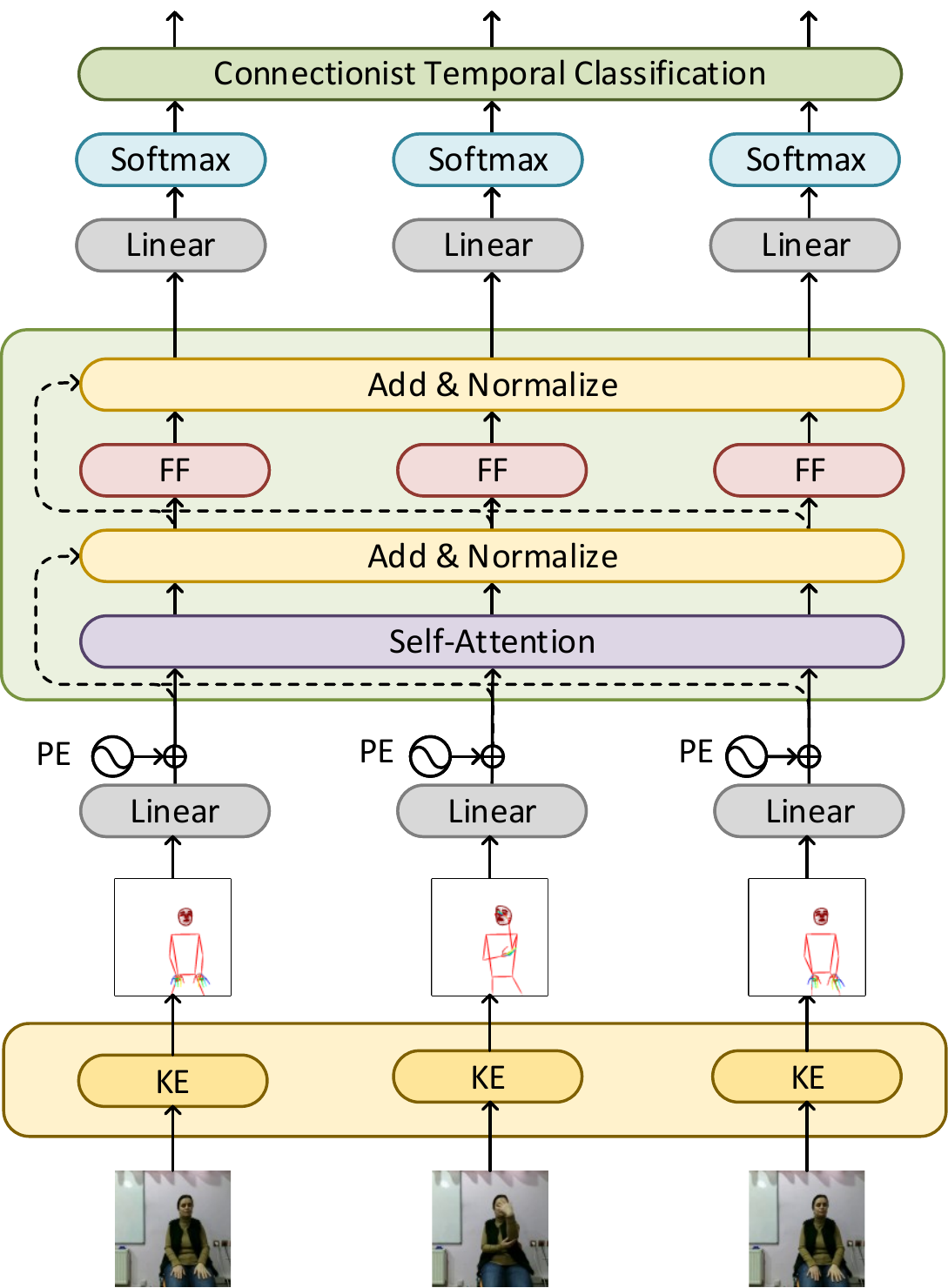}
    \caption{Diagram of \emph{Team 1}'s model with one subnetwork (in green). (KE: Keypoint extraction, PE: Positional encoding, FF: feed forward)}
    \label{fig:team-1-model}
\end{figure}

\emph{Team 1} worked with OpenPose \cite{pose:hidalgo2019singlenetwork} pose estimation output and used the SLR transformer architecture from Camgöz et al. \cite{camgoz2020sign}. The model takes as input a series of feature vectors, in this case, human upper body skeletal coordinates extracted from the video frames. These are each projected to a lower dimension hidden state vector. The size of the hidden state remains constant throughout the subsequent operations. A sinusoidal positional encoding is added to provide temporal information. This is then passed to a subnetwork consisting of a multiheaded self-attention layer, followed by a feedforward layer. After each of these layers, the output is added to the input and normalized. This subnetwork can be repeated any number of times. Finally, the output is fed to a linear layer and softmax to give probabilities for each class (Figure \ref{fig:team-1-model}).

The model is trained using CTC loss. This is designed to allow the output to be invariant to alignment; however, this is not a significant concern when there should only be one output symbol. The final prediction is obtained via CTC beam search decoding, collapsing multiple same class outputs into one. As the model is trained to predict a single class per video, it does not predict different classes within a sequence.

The number of layers, heads, hidden size, and dropout rate affect the model complexity. There is, therefore, a tradeoff between sufficient complexity to model the data and overfitting. 

Additionally, as a baseline, the pose estimation keypoints were replaced with the output of three off-the-shelf image-based frame feature extractors, giving us small dense representations for each frame. Three extractors were used: 1. EfficientNet-B7 \cite{tan2019efficientnet}; 2. I3D trained on Kinetics \cite{carreira2017quo}; and 3. I3D trained on BSL1K \cite{albanie20_bsl1k}.

\begin{figure}[b]
    \centering
    \includegraphics[width=\linewidth]{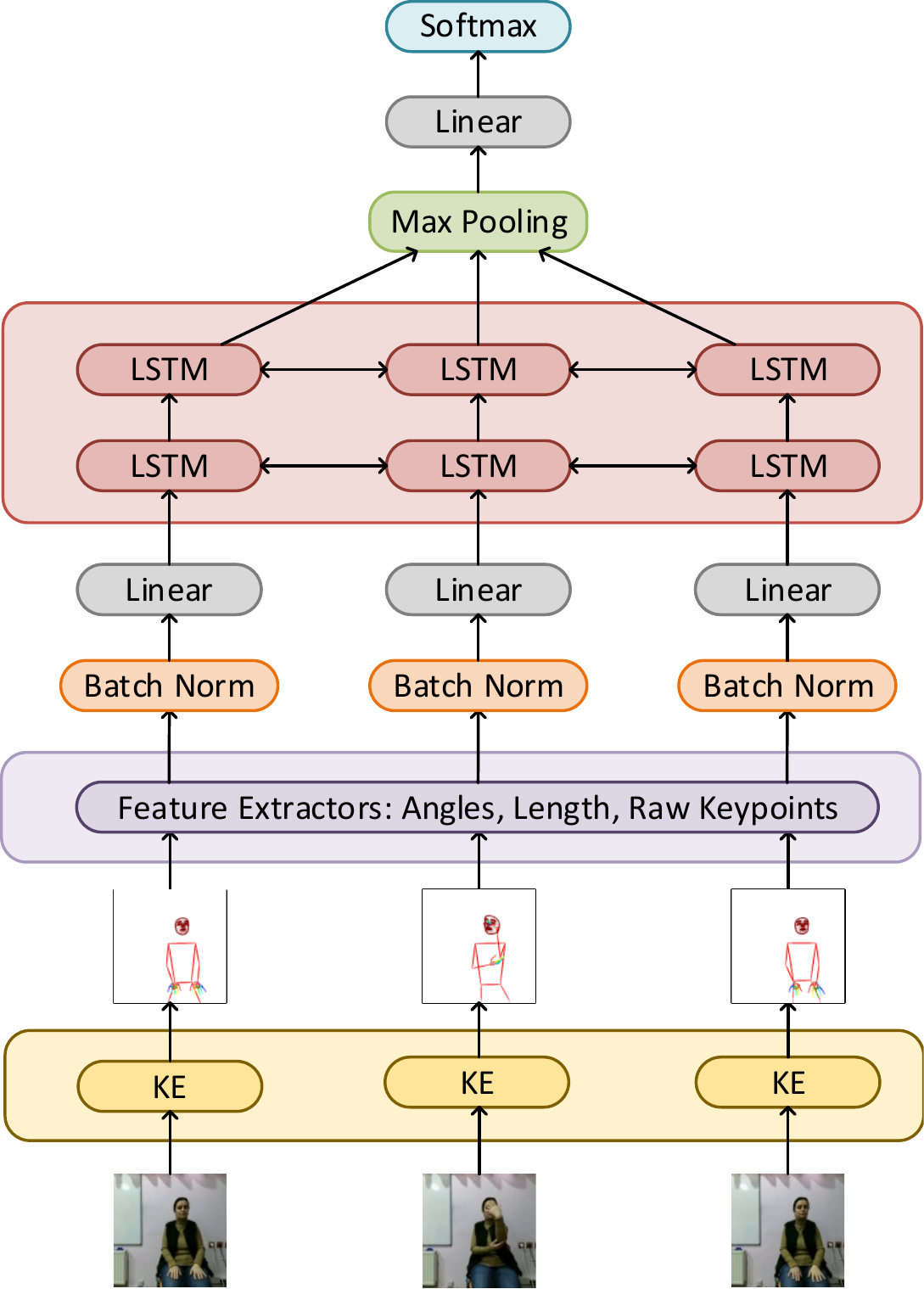}
    \caption{Diagram of \emph{Team 2}'s model. (KE: Keypoint extraction)}
    \label{fig:team-2-model}
\end{figure}

\subsection{Team 2}
\emph{Team 2} worked with the MediaPipe Holistic \cite{mediapipe2020holistic} pose estimation system output.
From the 543 landmarks, the face mesh was removed which consists of 468 landmarks and the remaining 75 landmarks were used for the body and hands.

A standard sequence classification architecture was used.
The model takes as input a series of feature vectors, constructed from a flat vector representation of the pose concatenated with the 2D angle and length of every limb, using the \emph{pose-format}\footnote{\url{https://github.com/AmitMY/pose-format}} library.
These representations are subjected to a 20\% dropout, normalized using 1D batch normalization, and are projected to a lower dimension hidden state vector (512 dimensions).
This is then passed to a two-layer BiLSTM with hidden dimension 256, followed by a max-pooling operation to obtain a single representation vector per video. 
Finally, the output is fed to a linear layer and softmax to give probabilities for each class (Figure \ref{fig:team-2-model}).

The model is trained using cross-entropy loss with the Adam optimizer (with default parameters) and a batch size of 512 on a single GPU. \emph{No} data augmentation or frame dropout is applied at training time, except for horizontal frame flip to account for left-handed signers in the dataset.

\section{Results}\label{sec:results}

Table \ref{table:results} shows our teams' results on the validation set. We note that both teams' approaches using pose estimation performed similarly, with validation accuracy ranging between 80\% and 85\%. It rules out trivial errors and implementation issues that, despite working independently, and with two separate pose estimation tools, both teams achieve similar evaluation scores.
Furthermore, from a comparison between the pose estimation based systems (80-85\%) and the pretrained image feature extractors (38-68\%), we can see that pose estimation features do indeed generalize better to the nature of the challenge, including unseen signers and backgrounds.

\begin{table}[h]
\centering
\begin{tabular}{lcc}
\toprule
& Team 1 & Team 2 \\
\midrule
EfficientNet-B7 & 38.80\% & ---  \\
I3D (Kinetics) & 47.46\% & ---  \\
I3D (BSL1K) & 68.65\% & --- \\ 
\midrule
OpenPose & 83.25\% & 79.99\% \\ 
Holistic & 85.63\% & 82.14\%  \\
\midrule
OpenPose+Holistic & 84.16\% & 82.89\% \\ \bottomrule

\end{tabular}
\caption{Results evaluated on the validation set with various frame-level features.}
\label{table:results}
\end{table}

We submitted \emph{Team 2}'s test set predictions to the official challenge evaluation.
On the test set, both \emph{OpenPose} and \emph{Holistic} performed \textbf{equally well} despite making different predictions, each with 78.35\% test set accuracy. However, our combined system, which was trained using both pose estimations, achieves 81.93\% test set accuracy.

\section{Analysis}\label{sec:analysis}

The interpretability of skeletal poses allows us to assess them qualitatively using visualisation.
We manually review our model's failure cases and categorize them into two main categories: hands interaction and hand-face interaction.

\begin{figure}[h]
    \centering
    \includegraphics[width=0.49\linewidth]{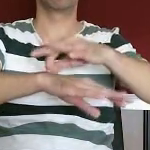}
    \includegraphics[width=0.49\linewidth]{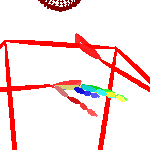}
    \caption{Example of hands interaction, where the pose estimation fails for one of the hands (Holistic).}
    \label{fig:hand-hand-1}
\end{figure}

\begin{figure}[h]
    \centering
    \includegraphics[width=0.49\linewidth]{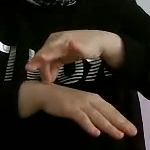}
    \includegraphics[width=0.49\linewidth]{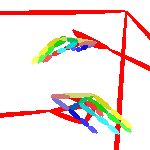}
    \caption{Example of hands interaction, where the pose estimation does not reflect the existing interaction (Holistic).}
    \label{fig:hand-hand-2}
\end{figure}

\paragraph{Hands Interaction}
When there exists an interaction between both hands, or one hand occludes the other from the camera's view, we often fail to estimate the pose of one of the hands (Figure \ref{fig:hand-hand-1}) or estimate it incorrectly such that the interaction is not clearly shown (Figure \ref{fig:hand-hand-2}).

\begin{figure}[h]
    \centering
    \includegraphics[width=0.49\linewidth]{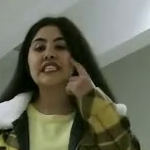}
    \includegraphics[width=0.49\linewidth]{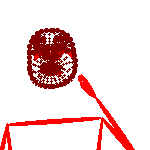}
    \caption{Example of hand-face interaction, where the pose estimation fails for the interacting hand (Holistic).}
    \label{fig:hand-face-1}
\end{figure}

\paragraph{Hand-Face Interaction}
When there exists an interaction between a hand and the face, or one hand overlaps with the face from the camera's angle, we often fail to estimate the pose of the interacting hand (Figure \ref{fig:hand-face-1}). 

These cases of missed interactions between the different body parts often lose the essence of the sign, where the interaction and the hand shape are the main distinguishing features for those signs, and thus hinder the model's ability to extract meaningful information from the pose that is relevant to the sign.

\begin{figure}[h]
    \centering
    \includegraphics[width=0.9\linewidth]{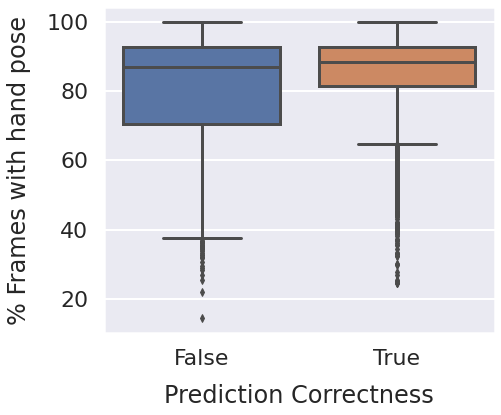}
    \caption{Distribution of percent of frames containing the Holistic pose estimation of the dominant hand in each validation sample, grouped by whether the final prediction of our model was correct.}
    \label{fig:sigtest}
\end{figure}

\paragraph{Presence or absence of hand pose} We describe a number of failure cases of Holistic pose estimation above. Many of them mean that keypoints for the hands are not available at all, since Holistic can omit them if it fails to detect the hand. As a complementary quantitative analysis, we correlate prediction outcomes with the average number of frames where hand pose was present (Figure \ref{fig:sigtest}).

We find that on average, for all correct predictions the percentage of frames that do contain hand keypoints (85.13\%) is significantly higher\footnote{We tested for a significant difference of the mean values with a Wilcoxon rank-sum test \cite{wilcoxon1992individual}, $p<0.0001$.} than for all incorrect predictions (79.78\%). This is in line with our qualitative analysis.

\section{Conclusions}\label{sec:conclusions}

Although many teams outperformed our models that use only off-the-shelf skeletal representations, with the best submission reaching 98.4\% test set accuracy, it is unclear how well such approaches will generalise to other datasets. Our initial questions related to how good skeletal representations are for recognition, given their natural ability to generalise. However, performance in the ChaLearn challenge suggests that despite their benefits, considerable information is lost in the skeletal representation that must be represented in the image domain. A qualitative analysis of our models' failure cases shows that pose estimation tools suffer from shortcomings when body parts interact. We conclude that pose estimation tools are not immediately applicable for the use in sign language recognition -- the current representations are not sufficiently expressive, and that further improvements with regard to interacting body parts is crucial for their applicability.

{\small
\bibliographystyle{ieee_fullname}
\bibliography{slp}

\begin{thebibliography}{10}\itemsep=-1pt

\bibitem{adaloglou2020comprehensive}
Nikolas Adaloglou, Theocharis Chatzis, Ilias Papastratis, Andreas Stergioulas,
  Georgios~Th Papadopoulos, Vassia Zacharopoulou, George~J Xydopoulos, Klimnis
  Atzakas, Dimitris Papazachariou, and Petros Daras.
\newblock A comprehensive study on sign language recognition methods.
\newblock {\em arXiv preprint arXiv:2007.12530}, 2020.

\bibitem{albanie20_bsl1k}
Samuel Albanie, G{\"u}l Varol, Liliane Momeni, Triantafyllos Afouras, Joon~Son
  Chung, Neil Fox, and Andrew Zisserman.
\newblock {BSL-1K}: {S}caling up co-articulated sign language recognition using
  mouthing cues.
\newblock In {\em ECCV}, 2020.

\bibitem{camgoz2017subunets}
Necati~Cihan Camg{\"o}z, Simon Hadfield, Oscar Koller, and Richard Bowden.
\newblock Subunets: End-to-end hand shape and continuous sign language
  recognition.
\newblock In {\em 2017 IEEE International Conference on Computer Vision
  (ICCV)}, pages 3075--3084. IEEE, 2017.

\bibitem{cihan2018neural}
Necati~Cihan Camg{\"o}z, Simon Hadfield, Oscar Koller, Hermann Ney, and Richard
  Bowden.
\newblock Neural sign language translation.
\newblock In {\em Proceedings of the IEEE Conference on Computer Vision and
  Pattern Recognition}, pages 7784--7793, 2018.

\bibitem{camgoz2020multi}
Necati~Cihan Camg{\"o}z, Oscar Koller, Simon Hadfield, and Richard Bowden.
\newblock Multi-channel transformers for multi-articulatory sign language
  translation.
\newblock In {\em European Conference on Computer Vision}, pages 301--319,
  2020.

\bibitem{camgoz2020sign}
Necati~Cihan Camg{\"o}z, Oscar Koller, Simon Hadfield, and Richard Bowden.
\newblock Sign language transformers: Joint end-to-end sign language
  recognition and translation.
\newblock In {\em Proceedings of the IEEE/CVF Conference on Computer Vision and
  Pattern Recognition}, pages 10023--10033, 2020.

\bibitem{pose:cao2018openpose}
Z. {Cao}, G. {Hidalgo Martinez}, T. {Simon}, S. {Wei}, and Y.~A. {Sheikh}.
\newblock Openpose: Realtime multi-person 2d pose estimation using part
  affinity fields.
\newblock {\em IEEE Transactions on Pattern Analysis and Machine Intelligence},
  2019.

\bibitem{pose:cao2017realtime}
Zhe Cao, Tomas Simon, Shih-En Wei, and Yaser Sheikh.
\newblock Realtime multi-person 2d pose estimation using part affinity fields.
\newblock In {\em CVPR}, 2017.

\bibitem{carreira2017quo}
Joao Carreira and Andrew Zisserman.
\newblock Quo vadis, action recognition.
\newblock {\em A new model and the kinetics dataset. CoRR, abs/1705.07750},
  2(3):1, 2017.

\bibitem{chai2013sign}
Xiujuan Chai, Guang Li, Yushun Lin, Zhihao Xu, Yili Tang, Xilin Chen, and Ming
  Zhou.
\newblock {Sign Language Recognition and Translation with Kinect}.
\newblock In {\em Proceedings of the International Conference on Automatic Face
  and Gesture Recognition (FG)}, 2013.

\bibitem{charles2014automatic}
James Charles, Tomas Pfister, Mark Everingham, and Andrew Zisserman.
\newblock {Automatic and Efficient Human Pose Estimation for Sign Language
  Videos}.
\newblock {\em International Journal of Computer Vision (IJCV)}, 110(1), 2014.

\bibitem{pose:chen2017adversarial}
Yu Chen, Chunhua Shen, Xiu-Shen Wei, Lingqiao Liu, and Jian Yang.
\newblock Adversarial posenet: A structure-aware convolutional network for
  human pose estimation.
\newblock In {\em Proceedings of the IEEE International Conference on Computer
  Vision}, pages 1212--1221, 2017.

\bibitem{cooper2010sign}
Helen Cooper and Richard Bowden.
\newblock {Sign Language Recognition using Linguistically Derived Sub-Units}.
\newblock In {\em Proceedings of the 4th Workshop on the Representation and
  Processing of Sign Languages: Corpora and Sign Language Technologies}, 2010.

\bibitem{cooper2011sign}
Helen Cooper, Brian Holt, and Richard Bowden.
\newblock {Sign Language Recognition}.
\newblock In {\em Visual Analysis of Humans}. Springer, 2011.

\bibitem{cui2019deep}
Runpeng Cui, Hu Liu, and Changshui Zhang.
\newblock A deep neural framework for continuous sign language recognition by
  iterative training.
\newblock {\em IEEE Transactions on Multimedia}, 21(7):1880--1891, 2019.

\bibitem{deng2009imagenet}
Jia Deng, Wei Dong, Richard Socher, Li-Jia Li, Kai Li, and Li Fei-Fei.
\newblock Imagenet: A large-scale hierarchical image database.
\newblock In {\em 2009 IEEE conference on computer vision and pattern
  recognition}, pages 248--255. Ieee, 2009.

\bibitem{fang2017deepasl}
Biyi Fang, Jillian Co, and Mi Zhang.
\newblock {DeepASL: Enabling Ubiquitous and Non-Intrusive Word and
  Sentence-Level Sign Language Translation}.
\newblock In {\em Proceedings of the ACM Conference on Embedded Networked
  Sensor Systems (SenSys)}, 2017.

\bibitem{graves2006connectionist}
Alex Graves, Santiago Fern{\'a}ndez, Faustino Gomez, and J{\"u}rgen
  Schmidhuber.
\newblock Connectionist temporal classification: labelling unsegmented sequence
  data with recurrent neural networks.
\newblock In {\em Proceedings of the 23rd international conference on Machine
  learning}, pages 369--376, 2006.

\bibitem{mediapipe2020holistic}
Ivan Grishchenko and Valentin Bazarevsky.
\newblock Mediapipe holistic.
\newblock
  \url{https://ai.googleblog.com/2020/12/mediapipe-holistic-simultaneous-face.html},
  2020.

\bibitem{pose:alp2018densepose}
R{\i}za~Alp G{\"u}ler, Natalia Neverova, and Iasonas Kokkinos.
\newblock Densepose: Dense human pose estimation in the wild.
\newblock In {\em Proceedings of the IEEE Conference on Computer Vision and
  Pattern Recognition}, pages 7297--7306, 2018.

\bibitem{pose:hidalgo2019singlenetwork}
Gines Hidalgo, Yaadhav Raaj, Haroon Idrees, Donglai Xiang, Hanbyul Joo, Tomas
  Simon, and Yaser Sheikh.
\newblock Single-network whole-body pose estimation.
\newblock In {\em ICCV}, 2019.

\bibitem{dataset:huang2018video}
Jie Huang, Wengang Zhou, Qilin Zhang, Houqiang Li, and Weiping Li.
\newblock Video-based sign language recognition without temporal segmentation.
\newblock In {\em Proceedings of the AAAI Conference on Artificial
  Intelligence}, volume~32, 2018.

\bibitem{ko2019neural}
Sang-Ki Ko, Chang~Jo Kim, Hyedong Jung, and Choongsang Cho.
\newblock {Neural Sign Language Translation based on Human Keypoint
  Estimation}.
\newblock {\em Applied Sciences}, 9(13), 2019.

\bibitem{krizhevsky2012imagenet}
Alex Krizhevsky, Ilya Sutskever, and Geoffrey~E Hinton.
\newblock Imagenet classification with deep convolutional neural networks.
\newblock In {\em Advances in neural information processing systems}, pages
  1097--1105, 2012.

\bibitem{luong2015effective}
Thang Luong, Hieu Pham, and Christopher~D. Manning.
\newblock Effective approaches to attention-based neural machine translation.
\newblock In {\em Proceedings of the 2015 Conference on Empirical Methods in
  Natural Language Processing}, pages 1412--1421, Lisbon, Portugal, Sept. 2015.
  Association for Computational Linguistics.

\bibitem{luzardo2013head}
Marcos Luzardo, Matti Karppa, Jorma Laaksonen, and Tommi Jantunen.
\newblock {Head Pose Estimation for Sign Language Video}.
\newblock {\em Image Analysis}, 2013.

\bibitem{moryossef2021datasets}
Amit Moryossef.
\newblock Sign language datasets.
\newblock \url{https://github.com/sign-language-processing/datasets}, 2021.

\bibitem{detection:moryossef2020real}
Amit Moryossef, Ioannis Tsochantaridis, Roee~Yosef Aharoni, Sarah Ebling, and
  Srini Narayanan.
\newblock Real-time sign-language detection using human pose estimation.
\newblock 2020.

\bibitem{pfister2012automatic}
Tomas Pfister, James Charles, Mark Everingham, and Andrew Zisserman.
\newblock {Automatic and Efficient Long Term Arm and Hand Tracking for
  Continuous Sign Language TV Broadcasts}.
\newblock In {\em Proceedings of the British Machine Vision Conference (BMVC)},
  2012.

\bibitem{pose:pishchulin2012articulated}
Leonid Pishchulin, Arjun Jain, Mykhaylo Andriluka, Thorsten Thorm~{\"a} hlen,
  and Bernt Schiele.
\newblock Articulated people detection and pose estimation: Reshaping the
  future.
\newblock In {\em 2012 IEEE Conference on Computer Vision and Pattern
  Recognition}, pages 3178--3185. IEEE, 2012.

\bibitem{Shotton2011a}
J. Shotton, Andrew Fitzgibbon, M. Cook, Toby Sharp, M. Finocchio, R. Moore, A.
  Kipman, and A. Blake.
\newblock {Real-time Human Pose Recognition in Parts from Single Depth Images}.
\newblock In {\em Proceedings of the IEEE Conference on Computer Vision and
  Pattern Recognition (CVPR)}, 2011.

\bibitem{pose:simon2017hand}
Tomas Simon, Hanbyul Joo, Iain Matthews, and Yaser Sheikh.
\newblock Hand keypoint detection in single images using multiview
  bootstrapping.
\newblock In {\em CVPR}, 2017.

\bibitem{Sincan:CVPRW:2021}
Ozge~Mercanoglu Sincan, Julio C.~S. {Jacques Junior}, Sergio Escalera, and
  Hacer~Yalim Keles.
\newblock Chalearn {LAP} large scale signer independent isolated sign language
  recognition challenge: Design, results and future research.
\newblock In {\em Proceedings of the IEEE/CVF Conference on Computer Vision and
  Pattern Recognition (CVPR) Workshops}, 2021.

\bibitem{dataset:sincan2020autsl}
Ozge~Mercanoglu Sincan and Hacer~Yalim Keles.
\newblock Autsl: A large scale multi-modal turkish sign language dataset and
  baseline methods.
\newblock {\em IEEE Access}, 8:181340--181355, 2020.

\bibitem{tan2019efficientnet}
Mingxing Tan and Quoc Le.
\newblock Efficientnet: Rethinking model scaling for convolutional neural
  networks.
\newblock In {\em International Conference on Machine Learning}, pages
  6105--6114. PMLR, 2019.

\bibitem{dataset:joze2018ms}
Hamid Vaezi~Joze and Oscar Koller.
\newblock Ms-asl: A large-scale data set and benchmark for understanding
  american sign language.
\newblock In {\em The British Machine Vision Conference (BMVC)}, September
  2019.

\bibitem{vaswani2017attention}
Ashish Vaswani, Noam Shazeer, Niki Parmar, Jakob Uszkoreit, Llion Jones,
  Aidan~N Gomez, {\L}ukasz Kaiser, and Illia Polosukhin.
\newblock Attention is all you need.
\newblock In {\em Advances in neural information processing systems}, pages
  5998--6008, 2017.

\bibitem{vogler2005analysis}
Christian Vogler and Siome Goldenstein.
\newblock Analysis of facial expressions in american sign language.
\newblock In {\em Proc, of the 3rd Int. Conf. on Universal Access in
  Human-Computer Interaction, Springer}, 2005.

\bibitem{dataset:von2007towards}
Ulrich Von~Agris and Karl-Friedrich Kraiss.
\newblock Towards a video corpus for signer-independent continuous sign
  language recognition.
\newblock {\em Gesture in Human-Computer Interaction and Simulation, Lisbon,
  Portugal, May}, 11, 2007.

\bibitem{pose:wei2016cpm}
Shih-En Wei, Varun Ramakrishna, Takeo Kanade, and Yaser Sheikh.
\newblock Convolutional pose machines.
\newblock In {\em CVPR}, 2016.

\bibitem{wei2016cpm}
Shih-En Wei, Varun Ramakrishna, Takeo Kanade, and Yaser Sheikh.
\newblock {Convolutional Pose Machines}.
\newblock In {\em Proceedings of the IEEE Conference on Computer Vision and
  Pattern Recognition (CVPR)}, 2016.

\bibitem{wilcoxon1992individual}
Frank Wilcoxon.
\newblock Individual comparisons by ranking methods.
\newblock In {\em Breakthroughs in statistics}, pages 196--202. Springer, 1992.

\bibitem{zafrulla2011american}
Zahoor Zafrulla, Helene Brashear, Thad Starner, Harley Hamilton, and Peter
  Presti.
\newblock {American Sign Language Recognition with the Kinect}.
\newblock In {\em Proceedings of the ACM International Conference on Multimodal
  Interfaces (ICMI)}, 2011.

\end{thebibliography}
}

\end{document}